\documentclass[10pt,twocolumn,letterpaper]{article}

\usepackage{iccv}
\usepackage{times}
\usepackage{epsfig}
\usepackage{graphicx}
\usepackage{amsmath}
\usepackage{amssymb}
\usepackage{xcolor}
\usepackage{algorithm}
\usepackage{algorithmic}
\usepackage{booktabs}
\usepackage{multirow}
\usepackage{verbatim}
\usepackage{color}
\usepackage{float}
\usepackage{enumitem}
\usepackage{booktabs}
\usepackage{pifont}
\usepackage{soul}
\usepackage{nicefrac} 
\usepackage{microtype}
\usepackage{multirow}
\usepackage{verbatim}
\usepackage{color}
\usepackage{float}
\usepackage{enumitem}
\usepackage{booktabs}
\usepackage{tabulary,multirow,overpic,xcolor}
\usepackage{bbm}
\usepackage{subfig}
\usepackage{graphicx}

\definecolor{citecolor}{RGB}{34,139,34}
\definecolor{citecolor2}{HTML}{0071bc}
\usepackage[pagebackref=true,breaklinks=true,letterpaper=true,colorlinks,
citecolor=citecolor2,bookmarks=false]{hyperref}

\newcommand{\app}{\raise.17ex\hbox{$\scriptstyle\sim$}}

\def\x{$\times$}
\newcolumntype{x}[1]{>{\centering\arraybackslash}p{#1pt}}

\newlength\savewidth\newcommand\shline{\noalign{\global\savewidth\arrayrulewidth
		\global\arrayrulewidth 1pt}\hline\noalign{\global\arrayrulewidth\savewidth}}
\newcommand{\tablestyle}[2]{\setlength{\tabcolsep}{#1}\renewcommand{\arraystretch}{#2}\centering\footnotesize}

\makeatletter\renewcommand\paragraph{\@startsection{paragraph}{4}{\z@}
	{.5em \@plus1ex \@minus.2ex}{-.5em}{\normalfont\normalsize\bfseries}}\makeatother

\newcommand{\tblref}[1]{Table~\ref{#1}}
\newcommand{\sref}[1]{\S\ref{#1}}

\newcommand{\pacc}[1]{{\bf \fontsize{7.5}{42}\selectfont \color{citecolor!80}~(#1)}}
\usepackage{stmaryrd}

\DeclareMathOperator*{\argmax}{arg\,max}

\newcommand{\cmark}{\ding{51}}%
\usepackage[breaklinks=true,bookmarks=false]{hyperref}

\iccvfinalcopy 

\def\iccvPaperID{***} 

\ificcvfinal\fi

\begin{document}

\def\OURS{Multiview  Pseudo-Labeling\xspace}
\def\ours{multiview  pseudo-labeling\xspace}
\def\oursshort{MvPL\xspace}
\def\OURSFULL{?\xspace}

\title{ Multiview Pseudo-Labeling for  Semi-supervised Learning from Video}

\author{
	Bo Xiong \qquad
	Haoqi Fan \qquad
	Kristen Grauman \qquad
	Christoph Feichtenhofer \vspace{.8em}\\
	Facebook AI Research (FAIR)
		\vspace{.6em}
}

\maketitle

\begin{abstract}
  We present a multiview  pseudo-labeling approach to video learning, a novel framework that uses complementary views in the form of appearance and motion information for semi-supervised learning in video. The complementary views help obtain more reliable ``pseudo-labels" on unlabeled video, to learn stronger video representations than from purely supervised data. 
  Though our method capitalizes on multiple views, it nonetheless trains a  model that is shared across appearance and motion input and thus, by design, incurs no additional computation overhead at inference time. On multiple video recognition datasets, our method substantially outperforms its supervised counterpart, and compares favorably to previous work on standard benchmarks in self-supervised video representation learning.
\end{abstract}

\vspace{5pt}
\section{Introduction}\label{sec:introduction}
\vspace{5pt}

3D convolutional neural networks (CNNs)~\cite{tran2015learning, carreira2017quo, tran2018closer, feichtenhofer2019slowfast} have shown steady progress for video recognition, and  
{particularly} human action classification, over recent years. This progress also came with a shift from traditionally small-scale datasets to large amounts of labeled data~\cite{kay2017kinetics,carreira2018short,carreira2019short} to learn strong spatiotemporal
feature representations. Notably, as 3D CNNs are data hungry, their performance has never been able to reach the level of hand-crafted features \cite{wang2013action} when trained `\textit{from-scratch}' on smaller scale datasets~\cite{simonyan2014two}. 

However, collecting a large-scale annotated video dataset~\cite{goyal2017something,carreira2019short} for the task at hand is expensive and tedious as it often involves designing and implementing annotation platforms at scale and hiring crowd workers to collect annotations. For example, a previous study~\cite{sigurdsson2016much} suggests it takes at least one dollar to annotate a {single} video with 157 human activities.
Furthermore, the expensive 
annotation process needs to be repeated for each task of interest or when the label space needs to be expanded. Finally, another dilemma that emerges with datasets collected from the web is that they are vanishing over time as users delete their uploads, and therefore need to be replenished in a recurring fashion~\cite{smaira2020short}.

The  {goal} of this {work} 
is semi-supervised learning in video to learn from both labeled and \emph{unlabeled} data, {thereby reducing} the amount of annotated data required for training. Scaling video models with unlabeled data is a 
{setting of high practical interest}, since 
collecting large amounts of unlabeled video data 
requires minimal human effort. Still, {thus far} this area has received far less attention than fully supervised learning from video. 

\begin{figure}[t!]
	\centering
	\renewcommand{\tabcolsep}{0pt}
	\includegraphics[width=1\columnwidth]{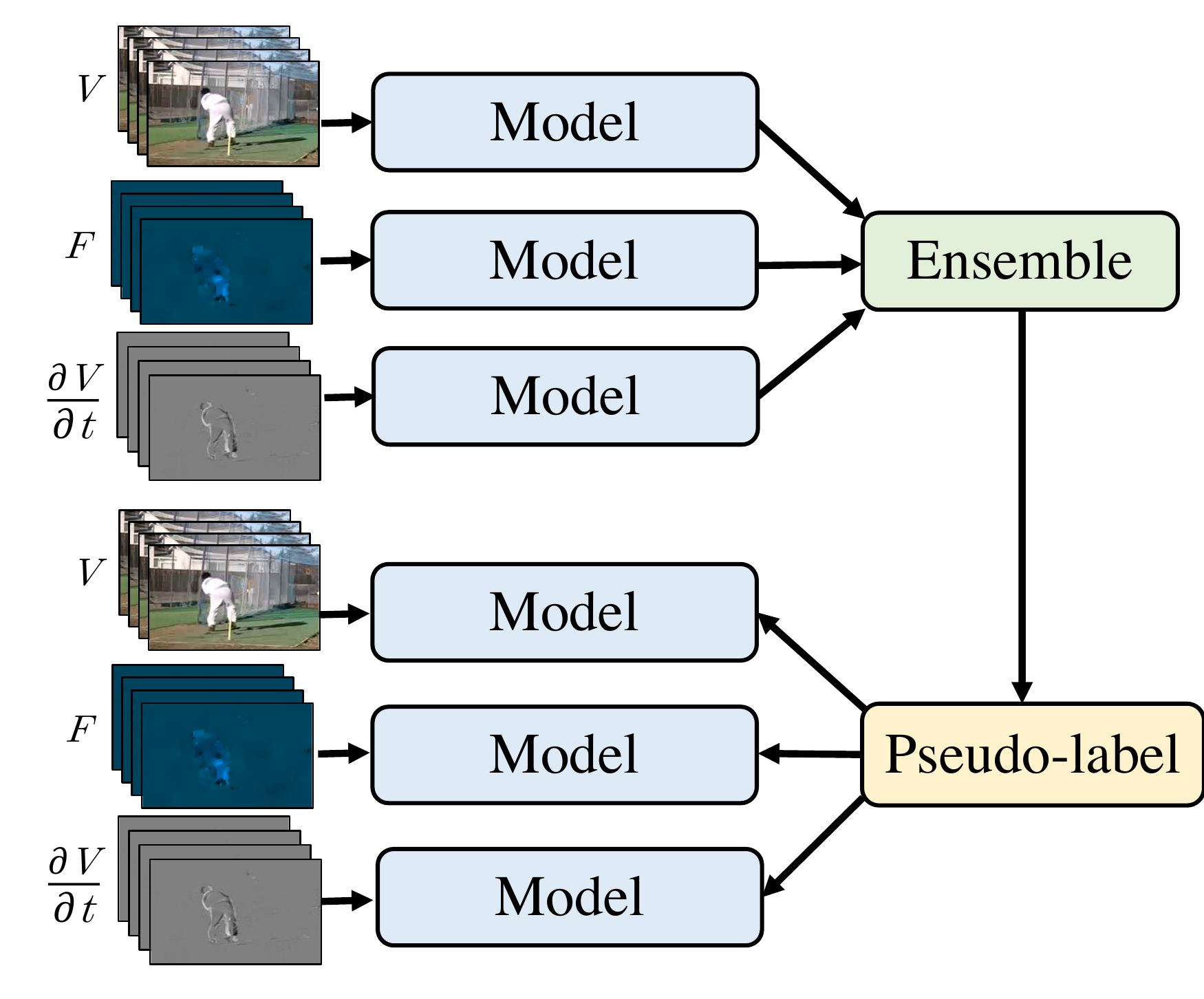}
	
	\caption{\textbf{Multiview  pseudo-labeling (\oursshort)} 
		takes in multiple complementary views of a single unlabeled video clip, in the form of RGB ($V$), optical-flow ($F$), and temporal gradients $(\frac{\partial V}{\partial t})$ {and uses a shared model} to perform semi-supervised learning. {After training, a single RGB view is used for inference.} }
	\label{fig:intro}
\end{figure}

Most prior advances in semi-supervised learning in computer vision focus on the problem of image  recognition. 
``Pseudo-labeling"~\cite{lee2013pseudo, yalniz2019billion, Xie_2020_CVPR, fixmatch} is a popular approach to utilize unlabeled images. The idea is  to use  the  predictions  from  a  model  as  target  labels and gradually add {the} unlabeled  {images (with their inferred labels)} to the training set. 
Compared to image recognition, {semi-supervised} video recognition presents its own challenges {and opportunities.  On the one hand,} 
the temporal dimension {introduces some} ambiguity, \ie given a video clip with an activity label, the activity may occur at any temporal location.
{On the other hand, video}
can also provide a valuable, complementary signal for recognition by the way objects move in space-time, \eg the actions `sit-down' \vs `stand-up' cannot be {discriminated} without using the temporal signal.  More specifically, {video adds} information about how actors, objects, and the 
{environment} \textit{change} over time. 

{Therefore,} directly applying semi-supervised learning algorithms designed for images to video could be sub-optimal (we will verify this point in our experiments), as image-based algorithms only consider appearance information and ignore the potentially rich dynamic structure captured by video. 

To address the challenge discussed above, we introduce \emph{multiview pseudo-labeling (\oursshort)}, a novel framework for semi-supervised learning {designed for} video. Unlike {traditional 3D CNNs} that implicitly learn spatiotemporal features from appearance,
our key idea is to {explicitly force a \emph{single model}} to learn appearance and motion {features} by {ingesting} multiple complementary \emph{views}\footnote{We use the term \emph{view} to refer to different input types (RGB frames, optical flow, or RGB temporal gradients), {as opposed to camera viewpoints.} in the form of  appearance, motion, and temporal gradients, {so as to} train the model from unlabeled data} {that augments labeled data.} 

We consider visual-only semi-supervised learning and all the views are computed from RGB frames.  
Therefore our method does not require any additional {modalities} nor {does it} require any change to the model architecture to {accommodate the additional views}.  
Our proposed multiview pseudo-labeling is general and can serve as a drop-in replacement for any pseudo-labeling based algorithm~\cite{lee2013pseudo, yalniz2019billion, Xie_2020_CVPR, fixmatch} that {currently} operates {only} on appearance, {namely} {by augmenting the model with multiple views and our ensembling approach to infer pseudo-labels.}

Our method
rests on two key technical insights: 1) a single model that nonetheless benefits from multiview data; and 2) an ensemble approach to infer pseudo-labels.  

First, we convert both optical flow  and temporal  difference to the same input format as RGB frames so that all the views can share the \emph{same} 3D CNN model.  The 3D CNN model takes only one view at a time and treats optical flow and temporal gradients as if they are RGB frames. The advantage 
is that we directly encode appearance and motion in the input space and distribute the information through multiple views, {to the benefit of the 3D CNN.}

Second,  when predicting pseudo-labels for unlabeled data, we use an ensemble of all the views for prediction. 
We show that predicting pseudo-labels from all the views {is} more effective than predicting from a single view alone. 
Our method uses a single model that can seamlessly accommodate different views as input for video recognition. 
See Figure~\ref{fig:intro} for an overview of our approach.

In summary, this paper makes the following contributions:
\begin{itemize}

	\item This work represents {an exploration in semi-supervised learning for video understanding,} 
	an area that is heavily researched in image understanding 
	~\cite{chen2020big, grandvalet2005semi, henaff2020data, fixmatch, uda}.
	{Our evaluation establishes semi-supervised baselines on Kinetics-400 (1\% and 10\% label case), and UCF101 (similarly as the image domain which uses 1\% and 10\% of labels on ImageNet~\cite{chen2020big, henaff2020data, fixmatch, uda}). 
	}

	\item Our technical contribution is a novel \ours framework for \textit{general} application in semi-supervised learning from video, that delivers consistent improvement in accuracy on \textit{multiple} pseudo-labeling algorithms. 
	
	\item On several challenging video recognition benchmarks, our method substantially improves {its single view} counterpart. We obtain state-of-the-art performance on UCF101~\cite{soomro2012ucf101} and HMDB-51 \cite{Kuehne11} when using Kinetics-400~\cite{kay2017kinetics} as unlabeled data, and outperform video self-supervised methods in this setting. 
	
\end{itemize}

\section{Related Work}\label{sec:related}

\paragraph{Semi-supervised learning in images.}
Most prior advances in semi-supervised learning in computer vision focus on 
image recognition. Regularization on unlabeled data is a common strategy {for semi-supervised learning}. Entropy regularization~\cite{grandvalet2005semi} minimizes the conditional entropy of class probabilities for unlabeled data. Consistency regularization forces the model representations to be similar when augmentations are applied to unlabeled data~\cite{sajjadi2016regularization}. 
VAT~\cite{miyato2018virtual} uses adversarial perturbations while UDA~\cite{uda} applies  RandAugment~\cite{randaug} for augmentations.
Pseudo-labeling~\cite{lee2013pseudo, yalniz2019billion} or self-training~\cite{Xie_2020_CVPR} is another common strategy for semi-supervised learning, 
{where} predictions from a model {are used} as pseudo-labels for unlabeled data. 
Pseudo-labels can be generated using a consensus from previous model
checkpoints~\cite{temporalensem} or an exponential moving average of model parameters~\cite{meanteacher}.
FixMatch~\cite{fixmatch} predicts pseudo-labels from weak augmentation to guide learning for strong augmentation generated from RandAugment~\cite{randaug}.  
{Unlike any of the prior work above, we consider video, and our method leverages multiple complementary views. 
}

\paragraph{Semi-supervised learning in videos.}
Compared to images, semi-supervised learning for video has received much less attention. 
The work of~\cite{zeng2017semi} applies an encoder-decoder framework to minimize a reconstruction loss.
The work in~\cite{videossl} combines pseudo-labeling and distillation~\cite{girdhar2019distinit} from a 2D image classifier to assist video recognition. 
However, {none of} the prior semi-supervised work capitalizes on the rich views (appearance, motion, and temporal gradients) in videos. To the best of our knowledge, we are the first to explore multiple 
complementary views for semi-supervised video recognition.
Co-training~\cite{blum1998combining} is a seminal work for semi-supervised learning with two views, first introduced for the web page classification problem.
{Co-training learns separate models for each view, whereas we share a single model for all views. }
{Our idea has the key advantage that} a single model can directly leverage the complementary sources of information from all the views. 
{Our experiments} demonstrate that our design  {outperforms co-training for this video learning setting.}

\paragraph{Self-supervised learning.} Another common direction to leverage unlabeled video data is self-supervised learning. Self-supervised learning first learns feature representations from a pretext task (e.g., 
audio video synchronization~\cite{korbar2018cooperative},  clustering~\cite{alwassel2019self}, clip order~\cite{xu2019self}, and instance discrimination~\cite{qian2020spatiotemporal} etc.), where the labels are generated from the data itself, and then fine-tunes the model on downstream tasks with labeled data. 
Self-supervised learning in video can leverage modalities by learning the correspondence between visual and audio cues~\cite{patrick2020multi} or video and text~\cite{zhu2020actbert, miech2020end}. Appearance and motion~\cite{Han20} can be used to boost performance in a contrastive learning framework or address domain adaptation~\cite{munro20multi}. 
Self-supervised training learns task-agnostic features, whereas semi-supervised learning is task-specific.
As suggested in~\cite{zeng2017semi}, semi-supervised learning can also leverage a self-supervised task as pre-training, {\ie the two ideas are not exclusive, as we will also show in results.}

\paragraph{Multi-modal video recognition.} 
{Supervised} video recognition can benefit from multi-modal inputs. Two-stream networks~\cite{simonyan2014two, feichtenhofer2016convolutional} leverage both appearance and motion. 
Temporal gradients~\cite{wang2016temporal,zhao2018recognize} can be used in parallel with appearance and motion to improve video recognition.
{Beyond} visual cues, audio signals~\cite{xiao2020audiovisual,kazakos2019epic} can also assist video recognition. We consider visual-only semi-supervised learning for video recognition. Like~\cite{wang2016temporal,zhao2018recognize}, we  use appearance, motion, and temporal gradients, {but unlike any of these prior models, our approach addresses semi-supervised learning.}

\section{\OURS (\oursshort)}\label{sec:approach}
We focus on semi-supervised learning for videos and our objective is to train a model by using both labeled and unlabeled data. 

Our main idea is to capitalize on the complementarity of appearance and motion views {for semi-supervised learning from video}. 
We first describe how we extract multiple views from video (\sref{sec:multiple views}), followed by how we use a single model to seamlessly accommodate all the views for multiview learning and how we obtain pseudo-labels with a multiview ensemble approach (\sref{sec:multiview_learning}). Subsequently, \sref{sec:multiview_instantiations} outlines three concrete instantiations of our approach and \sref{sec:details} provides implementation specifics.

\subsection{Multiple views of appearance and dynamics}\label{sec:multiple views}
{Many} video understanding methods only consider a single view (\ie, RGB frames), thereby possibly {failing to model} the rich dynamics in videos. 
Our goal is to use three complementary  views in the form of RGB frames, optical flow, and RGB temporal gradients to investigate this. Our motivation is that:

(i) RGB frames ($V$) record the static appearance at each time point but do not directly 
provide contextual information about object/scene motion. 

(ii) Optical flow ($F$) explicitly captures motion by describing the instantaneous image velocities in both horizontal and vertical axes.

(iii) Temporal gradients $(\frac{\partial V}{\partial t})$  between two consecutive RGB frames encode appearance change and correspond to dynamic information that deviates from a purely static scene. 
{Compared to optical flow, temporal gradients accentuate changes at boundaries of moving objects.}

Even though all three views are related to, and can be estimated from, each other by solving for optical-flow using the brightness constancy equation~\cite{horn1981determining},
\begin{equation}
	\nabla^\top V \cdot F + \frac{\partial V}{\partial t} = 0,
	\label{eq:ofce}
\end{equation}
with $\nabla \equiv (\frac{\partial}{\partial x}, \frac{\partial}{\partial y})^\top$, $F$ being the point-wise velocity vectors of the video brightness $V$ and $\frac{\partial V}{\partial t}$ the temporal gradients at a single position in space, $\mathbf{x}=(x,y)^\top$, and time $t$, we find empirically that all three views {expose} complementary sources of information about appearance and motion that are useful for video recognition. This finding is related to the complementarity of hand-crafted space-time descriptors that have been successful in the past 
(\eg histograms of space/time gradients~\cite{dalal2005histograms,Klaeser2008,DollarPETS05} and optical flow \cite{dalal2006human,wang2013action}).

\subsection{Learning a single model from multiple views} \label{sec:multiview_learning}

One way to accommodate multiple views for learning is to
train a separate model for each view and co-train their parameters~\cite{blum1998combining}. However, each view only implicitly interacts with other views through predictions on unlabeled data. 
Another alternative is to use multiple network streams~\cite{simonyan2014two, feichtenhofer2016convolutional}. However, here, the {number of} parameters of the model and the inference time grow roughly linearly with the number of streams, and during testing each stream has to be processed. 

{Instead,} we propose to train a \textit{single} model for all the complementary views by converting all the views to the same input format (\ie we train a single model $f$, and it can take any view as input). 
By sharing the same model, the complementary views can serve as additional data augmentations to learn stronger representations.
Compared to training separate models, our model can directly benefit from all the views instead of splitting knowledge between multiple models. 
Further, this technique does not incur any additional computation overhead after learning as only a single view is used for inference. 

Formally, given a collection of labeled video data  
$ \mathcal{X} = \big\{(x_i=[x_i^{1},...,x_i^{M}],  y_i)\big\} $ for  $i \in (1,\ldots,N_{l})$, 
where $y_i$ is the label for video instance $x_i$,  
$N_{l}$ is the total number of labeled videos, and $M$ is the total number of views,  and a collection of unlabeled video data  $\mathcal{U} = \big\{u_i =[u_i^{1},...,u_i^{M}]\big\}$ for   $i \in (1,\ldots,N_{u})$, our goal is to  is to learn a classifier $f(x_i)$ by leveraging both labeled and unlabeled data.

We use a supervised cross entropy loss $\ell_s$ for labeled data  and another cross entropy loss  $\ell_u$ for unlabeled data. 
For our training batches, we assume $N_{u} = \mu N_{l}$ where $\mu$ is a hyper-parameter that balances the ratio of labeled  and unlabeled samples $N_{l}$ and $N_{u}$, respectively.

\paragraph{Supervised loss.} 

For labeled data,
we extend the supervised cross-entropy loss $H$ to all the views on labeled data:
\begin{align}
	\label{eq:sup_loss}
	\ell_s = \frac{1}{N_l M}\sum_{i=1}^{N_l} \sum_{m=1}^{M} H(y_i,f(A(x_i^m))),
\end{align}
where $y_i$ is the label, and $A$ is a family of augmentations (\eg cropping, resizing) applied to input $x_i^m$ on view $m$.

\paragraph{Pseudo-label generation from multiple views.} 
For the unlabeled data, we use an ensembling approach to obtain pseudo-labels. 
Given an unlabeled video with a view $u_i^m$, let $s_i^m$ denote the pseudo-label class distribution, which is required because some of the instantiations we consider in the next section can filter out samples if the prediction is not confident. Then, the pseudo-label can be obtained by taking $\hat{s}_i^m\,{=}\,\argmax(s_i^m)$.
The model's class distribution prediction is $q_i^m = f(A(u_i^m))$, where $A$ again corresponds to the family of augmentations applied to input $u_i^m$, and the class with the highest probablity is $\hat{q}_i^m\,{=}\,\argmax(q_i^m)$.

We explore the following variants to obtain pseudo-labels $\hat{s}_i^m$ given the class distribution prediction from all the views.
\begin{enumerate}[label=\roman*]
	\item\label{itm:first}  \textbf{Self-supervision.} For each  $u_i^m$, we directly use the most confident prediction $\hat{q}_i^m$ as its pseudo-label, that is, $\hat{s}_i^m=\hat{q}_i^m$.
	This is the most straightforward way to generate pseudo-labels. However, each view only supervises itself and does not benefit from the predictions from other views.

	\item\label{itm:second}  \textbf{Random-supervision.} 
	For each  $u_i^m$, we randomly pick another view 
	$n \in (1,\ldots,M)$ and use the prediction on that view as the pseudo-label. Then we have $\hat{s}_i^m=\hat{q}_i^n$.
	
	\item\label{itm:third}  \textbf{Cross-supervision.} We first build a bijection function $b(m)$ for each view such that each view is deterministically mapped to another view and does not map to itself.  Then for each $u_i^m$, we  have $\hat{s}_i^m=\hat{q}_i^{b(m)}$.
	This is similar to co-training~\cite{blum1998combining} in the two-view case.
	
	\item\label{itm:fourth}  \textbf{Aggregated-supervision.} For each unlabeled video, we obtain pseudo-labels by taking the weighted average of predictions from each view. 
	\begin{equation}
		\label{eq:loss1}
		s_i^m = \frac{1}{\sum_{m=1}^{M} {w_m}}\sum_{m=1}^{M} w_m f(A(u_{i}^{m})).
	\end{equation}
	Then we obtain the  pseudo-label by $\hat{s}_i^m\,{=}\,\argmax(s_i^m)$.
	Note that in this case, all the views from the video $u_{i}$ share the same pseudo-label. The advantage of this approach is that the pseudo-label contains information from all the views. {We specify how to obtain the weight for each view in the implementation details.}
\end{enumerate}

\paragraph{Unsupervised loss.} 
After obtaining the class distribution of each view  $q_i^m = f(A(u_i^m))$ {for} an unlabeled video with $M$ views ($u_i=[u_i^1,...,u_i^M]$), we use one of the variants (\ref{itm:first}) $-$ (\ref{itm:fourth}) to obtain pseudo-label class distribution $s_i^m$ and pseudo-label $\hat{s}_i^m$. 

The pseudo-label $\hat{s}_i^m$ is then used as training signal for learning from the same, but differently augmented, data $\hat{A}(u_i^m)$, where $\hat{A}(x)$ denotes another family of transformations applied to the same unlabeled video $u_i$. Then our unsupervised loss is 

\begin{equation}
	\label{eq:unsup_loss}
	\ell_u = \frac{1}{\mu N_l M}\sum_{i=1}^{\mu N_l} \sum_{m=1}^{M} \mathbbm{1}  (\max(s_i^m\ge \tau)), H(\hat{s}_i^m,f(\hat{A}(u_i^m))
\end{equation}
where $\tau$ is a threshold used to filter out unlabeled data if the prediction is not confident. The total loss is $\ell = \ell_l  + \lambda_u \ell_u,$ where $\lambda_u$ controls the weight for the unlabeled data. {Sec.~\ref{sec:details} provides implementation details on the specific augmentations used.}

\subsection{\oursshort instantiations} \label{sec:multiview_instantiations}
 Our \oursshort framework is generally applicable to multiple semi-supervised learning algorithms that are based on pseudo-labels.
We instantiate our approach by unifying multiple methods in the same framework and analyze the commonality across methods. In this paper we concentrate on Pseudo-Label~\cite{lee2013pseudo},  FixMatch~\cite{fixmatch}, and UDA~\cite{uda}. On a high-level, these methods only differ in their utilization of unsupervised data in eq. \eqref{eq:unsup_loss}. We summarize our instantiations next.

\paragraph{Pseudo-Label.} Pseudo-Label~\cite{lee2013pseudo} uses the prediction from a sample itself as supervision.
 To apply our framework with Pseudo-Label~\cite{lee2013pseudo}, {we} simply use the same family of augmentations for obtaining pseudo-labels and learning from pseudo-labels, \ie $\hat{A}=A$.

\paragraph{FixMatch.} The main idea for FixMatch~\cite{fixmatch} is to predict pseudo-labels from weakly-augmented data and then use the pseudo-label as the learning target for a strongly-augmented version of the same data.
{Given an unlabeled {image},}
weakly-augmented data is obtained by applying standard data augmentation strategies, $A$,
that include flipping and cropping.
Strongly-augmented data is obtained by applying a family of augmentation operations $\hat{A}$, such as rotation, contrast, and sharpness \etc, using RandAugment~\cite{randaug}, that significantly alter the appearance of the unlabeled data.

\paragraph{Unsupervised Data Augmentation (UDA).}
{Similar to FixMatch~\cite{fixmatch}, UDA~\cite{uda} also uses weak and strong augmentations by enforcing consistency between them in forms of predicted class distributions. To extend UDA with \oursshort, we}
first sharpen the predicted class distribution $s_i^m$  to obtain \mbox{$\tilde{s}_i^m$. We then replace the hard label $\hat{s}_i^m$ in Eq.~\ref{eq:unsup_loss} with 
$\tilde{s}_i^m$.}
 Strictly speaking, UDA~\cite{uda} is not a pseudo-labeling algorithm per-se, because it uses soft labels (predicted class distribution with sharpening) as the learning signal. 
 
\begin{figure}[t!]
	\centering
	\renewcommand{\tabcolsep}{0pt}
	\includegraphics[width=1\columnwidth]{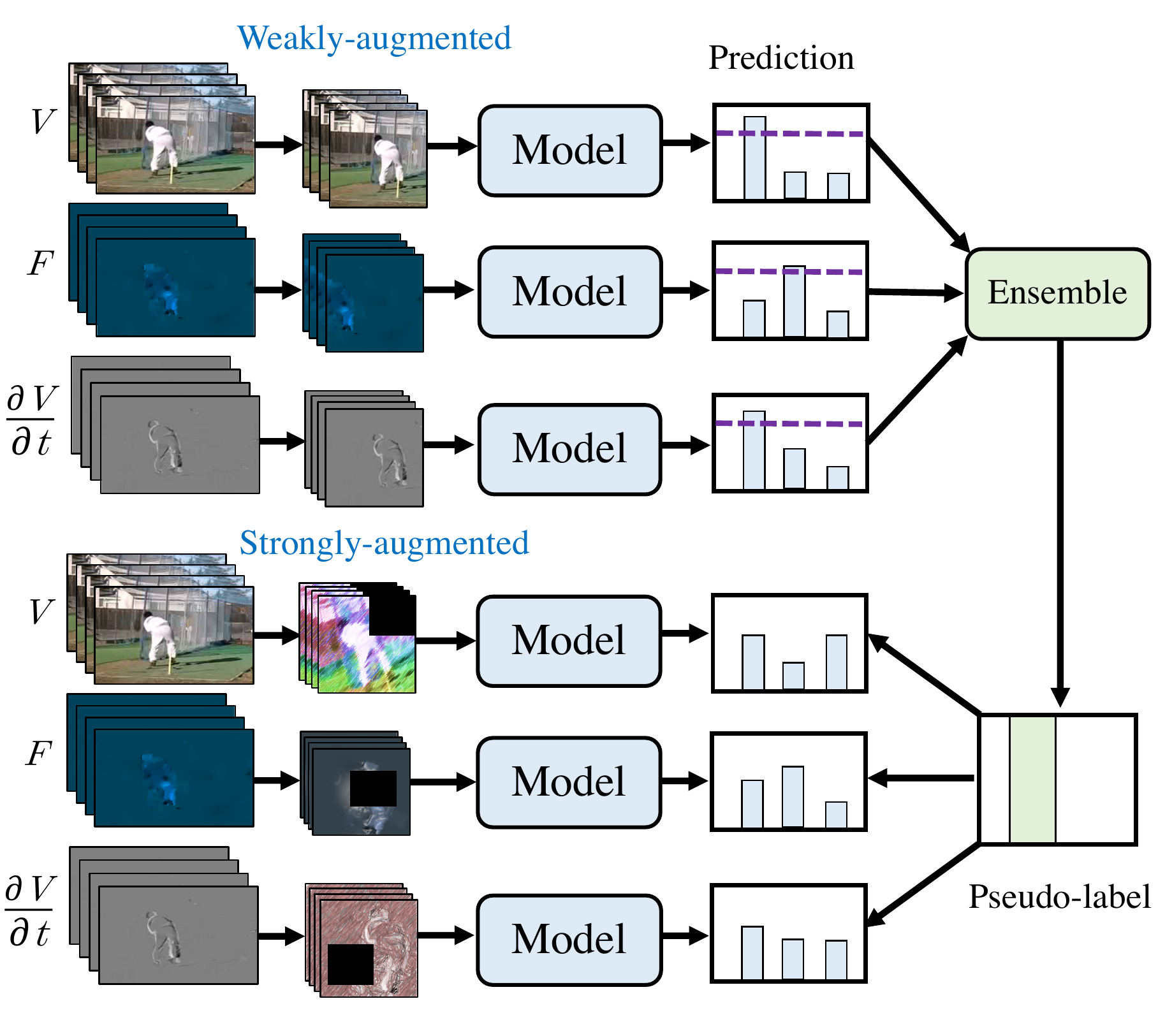}
	\caption{\textbf{Illustration of \oursshort} 
			applied to strongly-augmented data. Given an unlabeled video, we first obtain a weakly-augmented version of each view and then obtain predictions on them. Then we generate pseudo-labels by aggregating predictions from all the views. The pseudo-labels are used as a supervision signal for the 
		strongly-augmented version of each view from the same video.}
	\label{fig:app_fixmatch}
\end{figure}

We show an illustration of how to apply our method with strong augmentations in Figure~\ref{fig:app_fixmatch}.

\subsection{Implementation Details}\label{sec:details}

\paragraph{Model network architecture.} 
 As a backbone we use: R-50 \cite{He16} following the Slow pathway in \cite{feichtenhofer2019slowfast} with clips of $T{=}$8 frames sampled with stride $\tau{=}$8 from 64 raw-frames of video. This is a 3D ResNet-50 \cite{He16} without temporal pooling in the convolutional features. The input to the network is a clip of 8 frames with a sampling stride of 8, covering 64 frames of the raw video. The spatial input size is $224 \times 224$. 

\paragraph{Inference.} 
We follow the test protocol in~\cite{feichtenhofer2019slowfast}. The video model only takes RGB frames as input at inference {time}.  
For each video, we uniformly sample 10 clips along its temporal dimension. For each
clip, we scale the shorter spatial side to 256 pixels and take
3 crops of 256$\times$256. Finally, we obtain the prediction by averaging the softmax scores.

\paragraph{Converting optical flow and temporal gradients.} 
We pre-compute (unsupervised) optical flow using the software package of~\cite{liu2009beyond} that implements {a coarse-to-fine algorithm}~\cite{bruhn2005lucas}. We convert both the raw optical flow and RGB temporal gradients into 3-channel inputs that are in the same range as RGB frames. For optical flow, the 
{first two} channels correspond to displacements in the horizontal and vertical directions, respectively. The
{third} channel corresponds to the magnitude of the flow. All three channels are then normalized to the range of 0 and 255. 
We obtain temporal gradients by subtracting the next RGB frame from the current RGB frame. We then normalize them to the RGB range by adding 255 and dividing by 2.

\begin{table*}[t]\centering\vspace{-3mm}
	\subfloat[\textbf{\oursshort with different pseudo-labeling algorithms.} Our {semi-supervised} method consistently improves all three algorithms. \label{tab:general}]{
		\tablestyle{2pt}{1.02}
		\small
		\begin{tabular}{c|x{42}|x{42}|x{50}}
			\multirow{2}{*}{method} &Pseudo-Label~\cite{lee2013pseudo} & 	 \multirow{2}{*}{UDA~\cite{uda}} & \multirow{2}{*}{FixMatch~\cite{fixmatch}}  \\\shline
			base	& 30.4 & 47.0 & 48.5 \\
			\textbf{\oursshort} 	 & \textbf{70.0}  \pacc{+39.6}	 & \textbf{77.8} \pacc{+30.8}	& \textbf{79.1} \pacc{+30.6}	\\ 
			\multicolumn{3}{c}{~}\\
			\multicolumn{3}{c}{~}\\
			\multicolumn{3}{c}{~}\\
		\end{tabular}
	}\hspace{20pt}
	\subfloat[\textbf{Complementarity of views.} 
	Optical Flow and temporal gradients (TG) are complementary to RGB.  \label{tab:ab_view}]{
		\tablestyle{2pt}{1.02}
		\small
		\begin{tabular}{c|c|c|c}
			RGB & Flow & TG &   \textbf{\oursshort}	\\
			\shline
			\cmark &  &  & 48.5\\ 
			\shline
			\cmark & \cmark &  &  76.5 \pacc{+28.0}\\ 
			\cmark &  & \cmark &  74.0 \pacc{+25.5}\\ 
			\cmark & \cmark &  \cmark&  \textbf{79.1} \pacc{+30.6}\\ 
			\multicolumn{4}{c}{~}\\
			\multicolumn{4}{c}{~}\\
		\end{tabular}
	}\hspace{20pt}
	\subfloat[\textbf{Ways to generate  pseudo-labels}. Aggregated-supervision incroporating (All) views obtains the best result. \label{tab:ab_model}]{
		\tablestyle{2pt}{1.02}
		\small
		\begin{tabular}{l|x{20}}
			\multicolumn{1}{c|}{Supervision}    & Top 1 	\\ \shline
			Self (\ref{itm:first}) & 75.8	\\
			Random (\ref{itm:second})& 75.5	\\
			Cross (\ref{itm:third}) 1) & 78.7	\\
			Cross (\ref{itm:third}) 2) & 76.8	\\
			Agg. (\ref{itm:fourth})  (Exclusion)  &78.1	\\
			Agg. (\ref{itm:fourth})  (All)  &\textbf{79.1}	\\ 
		\end{tabular}
	}
	\vspace{-5pt}
	
	\caption{\textbf{Ablation study} on UCF101 split-1. We use \textit{only 10\%} of its training labels and the \textit{entire} training set as unlabeled data. We report top-1 accuracy on the validation set. Backbone: R-50, Slow-pathway~\cite{feichtenhofer2019slowfast}, $T\times\tau=$ 8\x 8. }
	\vspace{-5pt}
	\label{tab:ablations}
\end{table*}


\paragraph{Video augmentations.} For weak augmentation, we use default video classification augmentations~\cite{feichtenhofer2019slowfast}. In particular, given a video clip, we first randomly flip it horizontally with a 50\% probability, and then we crop 224$\times$224 pixels from the video clip with a shorter side randomly sampled between 256 and 320 pixels.

As strong augmentations, we apply RandAugment~\cite{randaug} followed by Cutout~\cite{cutout} (we randomly cut a 128$\times$128 patch from the same location across all frames in a video clip). RandAugment~\cite{randaug} includes a collection of image transformation operations (e.g., rotation, color inversion, translation, contrast adjustment, etc.). It randomly selects a small set of transformations to apply to data. 
RandAugment~\cite{randaug} contains a hyperparameter  that controls the severity of all operations. We follow a random magnitude from 1 to 10 at each training step.
When applying RandAugment to video clips, we keep the spatial transformations temporally consistent across all frames in a video clip.

\paragraph{Curriculum learning.} We find it useful to first warm up the training in the first few epochs with only the labeled data and then start training with both labeled and unlabeled data.

\paragraph{Training details.}
We implement our model with \mbox{PySlowFast}~\cite{fan2020pyslowfast}.
We adopt synchronized SGD training in 64 GPUs following the recipe in \cite{Goyal2017}, and we found its accuracy is as good as typical training in one 8-GPU machine. 
We follow the learning rate schedule used in~\cite{feichtenhofer2019slowfast}, which combines 
a half-period cosine schedule \cite{Loshchilov2016} of learning rate decaying and a 
linear warm-up strategy \cite{Goyal2017}.
We use momentum of 0.9 and weight decay of 10$^\text{-4}$. Dropout \cite{srivastava2014dropout} of 0.5 is used before the final classifier layer.
Please see supp. for additional details.
For \oursshort with FixMatch, we set the threshold $\tau$ to 0.3 (used for filtering training samples if the prediction is not confident). We set the ratio $\mu$ (a ratio to balance the number of labeled and unlabeled data) to 3 for
Kinetics-400~\cite{kay2017kinetics} and set $\mu$ to 4 for
UCF101 \cite{soomro2012ucf101} and HMDB51~\cite{kuehne2011hmdb}. For aggregated-supervision (\ref{itm:fourth}), we assign each view with the same weight $w_m$ as we found this works well in practice. \sref{sec:app_details} provides further specifics.

\section{Experiments}\label{sec:results_main}

We validate our approach for semi-supervised learning for video.
First, we present ablation studies to validate our design choices  in  Sec.~\ref{sec:ablation}.
Then, we show the main results by evaluating our proposed method on multiple video recognition datasets in 
Sec.~\ref{sec:result_main}. 
{Finally, we  compare our method with existing self-supervised methods in Sec.~\ref{sec:out}.} 
Unless specified otherwise, we present results on our method used in conjunction with FixMatch~\cite{fixmatch} {using} aggregated- supervision to obtain pseudo-labels.

\paragraph{Datasets.} 
We evaluate our approach on three {standard} video recognition datasets: Kinetics-400~\cite{kay2017kinetics} (K400), UCF101~\cite{soomro2012ucf101} and HMDB51~\cite{kuehne2011hmdb}. K400 contains 400 action classes with roughly 240k training videos. Each video is around 10 seconds. UCF101 contains 101 action classes with roughly 10k training videos and HMDB51 contains 51 action classes with roughly 4k training videos.
Both UCF101 and HMDB51 have 3 train-val splits.

\subsection{Ablation Studies}\label{sec:ablation}
We first investigate ablation studies to examine the effectiveness of \oursshort.
Our ablations are carried out on UCF101 split-1 and use \textit{only 10\%} of its training labels and the \textit{entire} UCF101 training set as unlabeled data (evaluation is done on the validation set).  
For all {ablation} experiments, 
we train the network for 600 epochs from scratch {with no warm-up} and use \textit{aggregated-supervision} (\ref{itm:fourth}) from all the views to obtain pseudo-labels, unless specified otherwise.

\paragraph{\oursshort generally improves pseudo-labeling techniques.
}

Table~\ref{tab:general} studies the effect of instantiating \oursshort with various pseudo-labeling algorithms, as outlined in \sref{sec:multiview_instantiations}.
\oursshort consistently improves all three algorithms by a large margin with an average absolute gain of 33.7\%. Pseudo-Label~\cite{lee2013pseudo} receives a larger gain (+39.6), presumably as it only relies on weak augmentations, while  UDA~\cite{uda}, and FixMatch~\cite{fixmatch} are using strong augmentations (RandAugment~\cite{randaug}) that lead to higher baseline performance for these methods. 

The results show that the \oursshort  framework provides a general improvement for \emph{multiple} baselines, instead of only improving  \emph{one} baseline. {This suggests} that \oursshort  { is} not tied to any particular pseudo-labeling algorithm and can be used to \textit{generally} {to} enhance existing pseudo-labeling algorithms for video understanding. Since FixMatch provides slightly higher performance than UDA, we 
use it for all subsequent experiments.

\paragraph{{Complementarity of views.}}
We now study how the different views contribute to the performance. We report results in Table~\ref{tab:ab_view} for using \oursshort on the FixMatch baseline from Table~\ref{tab:general}, with different views added one-by-one.
With RGB input alone, the FixMatch model fails to learn a strong representation (48.5\%). Adding complementary views that encode motion information immediately boosts performance. 
We observe an absolute gain of \textbf{+28.0}\% when additionally using Flow to RGB frames, and a \textbf{+25.5}\% gain for using temporal gradients (TG). The last row in Table~\ref{tab:ab_view} shows that both optical flow and temporal gradients are complementary to RGB frames as adding both views can significantly improve performance by \textbf{+30.6}\%. Here, it is important to note that \textit{test-time computation cost of all these variants is identical}, since \oursshort only uses the additional views during training.

\paragraph{Impact of pseudo-label variants.}
Table~\ref{tab:ab_model} explores different ways to generate pseudo-labels, namely, self-supervision (\ref{itm:first}), random-supervision (\ref{itm:second}), cross-supervision (\ref{itm:third}) and aggregated-supervision (\ref{itm:fourth}).

We make the following observations: \vspace{2pt}

Self-supervision (\ref{itm:first}) and random-supervision (\ref{itm:second}) obtain relatively low performance. This could be because self-supervision only bootstraps from its own view and does not take full advantage of other complementary views, and  random-supervision randomly picks a view to generate pseudo-labels, which we hypothesize might hinder the learning process, as the  learning targets change stochastically.

For cross-supervision (\ref{itm:third}), we show two variants with different bijections\footnote{
The notation A $\Leftarrow$ B indicates view A uses the pseudo-labels predicted from view B. }: 

1) RGB  $\Leftarrow$ Flow,  
Flow  $\Leftarrow$ TG, 
TG  $\Leftarrow$ RGB;

2) RGB  $\Leftarrow$ TG,  
Flow  $\Leftarrow$ RGB, 
TG  $\Leftarrow$ Flow.

Both variants of cross-supervision obtain better performance than self-supervision because both optical flow and temporal gradients are complementary to RGB frames and boost overall model accuracy. \vspace{2pt} 

\noindent For aggregated-supervision (\ref{itm:fourth}) we  examine two variants:  

(Exclusion): weighted average excluding self view; 

 (All): weighted average from all the views.

The Exclusion variant that uses aggregated-supervision from all the views obtains the best result with \textbf{79.1}\%. 
Here, we hypothesize that predictions obtained by an ensemble of all the views are more reliable than the prediction from any single view which leads to more accurate models.

\paragraph{Curriculum warm-up schedule.}
{Table} \ref{tab:ab_warm} shows an extra ablation on UCF101 {with 10\% labeled data}, described next.

\begin{table}[h!]
	\centering
	\tablestyle{8pt}{1.3}
	\small
	\begin{tabular}{cccccc}
		Warm-up epochs & 0 & 20 & 40 & 80 & 160  \\
		\shline
		Top-1 &79.1 & 79.5 & 80.4 & \textbf{80.5} & 80.3   \\
	\end{tabular}
	\caption{{Accuracy on UCF101 with 10\% labels used and a varying {supervised} warm-up duration.} {Supervised} warm-up with 80 epochs obtains the best results. }
	\label{tab:ab_warm} 
\end{table}

Before semi-supervised training, we employ a supervised warm-up that performs {training with only }the labeled data. Here, we compare the performance for different warm-up durations in our 10\% UCF-101 setting, \ie the same setting as in \tblref{tab:ablations} in which we train on UCF101 split-1, and use 10\% of its labeled data and the entire UCF101 training set as unlabeled data.

Table~\ref{tab:ab_warm} shows the results. 
Warm-up with 80 epochs obtains the best result of 80.5\% accuracy, 1.3\% better than not using supervised warm-up. We hypothesize that the warm-up allows the semi-supervised approach to learn with more accurate pseudo-label information in early stages of training.   
If the warm-up  is longer, the accuracy slightly degrades, possibly because the model converges to the labeled data early, and therefore is not able to fully use the unlabeled data for semi-supervised training.

\subsection{Results on Kinetics-400 and UCF-101}\label{sec:result_main}
\vspace{10pt}

We next evaluate our approach for semi-supervised learning on Kinetics-400, in addition to UCF-101. 
We 
consider two settings where 
1\% or 10\% of the labeled data are used. Again, the entire training dataset is used as unlabeled data. 
For K400, we form two balanced labeled subsets by sampling 6 and 60 videos per class. 
For UCF101, we use split 1 and sample 1 and 10 videos per class as labeled data.
Evaluation is again performed on the validation sets of K400 and UCF101.

\begin{table}[t]
	\centering
	\tablestyle{2.5pt}{1.1}
	\small
	\begin{tabular}{l|ll|ll}
		& \multicolumn{2}{c|}{Kinetics-400} & \multicolumn{2}{c}{UCF101}  \\
		\shline
		Method & 1\%  & 10\%    & 1\%  & 10 \%   \\
		\shline
		Supervised & 5.2 & 39.2 & 6.2 & 31.9  \\
		\textbf{\oursshort}  & \bf{17.0}\pacc{+11.8}	 & \bf{58.2}\pacc{+19.0} & \bf{22.8}\pacc{+16.6}  & \bf{80.5}\pacc{+48.6} 
	\end{tabular}
	\caption{\textbf{Results on K400 and UCF101}  when 1\% and 10\% of the labels are used for training. Our \oursshort  substantially outperforms the {direct} counterpart of supervised learning. Backbone: R-50, Slow-pathway~\cite{feichtenhofer2019slowfast}, $T\times\tau=$ 8\x 8.}
	\label{tab:main_results} 
\end{table}

We compare our semi-supervised \oursshort with the direct counterpart that uses supervised training on labeled data.
Table~\ref{tab:main_results} {shows the results}. We first look at the results in the supervised setting. With RGB input alone, the video model fails to learn a strong representation from limited labeled data: the model obtains 5.2\% and 6.2\% accuracy on K400 and UCF101 respectively when using 1\% of the labeled data, and 39.2\% and 31.9\% on K400 and UCF101 when using 10\% of the labels in the training data. 
 
On Kinetics, compared to the fully supervised approach, our semi-supervised \oursshort has an absolute gain of \textbf{+11.8}\% and \textbf{+19.0}\% when using 1\% and 10\% labels  respectively. This substantial improvement comes without cost at test time, as, again, only RGB frames are used for \oursshort inference. 
The gain in UCF101 is even more significant. {Overall, the results show that \oursshort can effectively learn a strong representation from unlabeled video data}.

\vspace{10pt}
\subsection{Comparison with self-supervised learning}~\label{sec:out}

{In a final experiment, we consider comparisons with self-supervised learning.}
Here, we evaluate our approach by using UCF101 and HMDB51 as the labeled dataset and Kinetics-400 as unlabeled data. 
{For both UCF101 and HMDB51,
we train and test on all three splits and report the average performance.}
 This setting is also common for self-supervised learning methods that are pre-tained on K400 and fine-tuned on UCF101 or HMDB51. We compare with the state-of-the-art approaches~\cite{alwassel2019self,avid,patrick2020multi,benaim2020speednet,Han20,yang2020video}.

\begin{table*}[t] 
	\centering
	\tablestyle{2.7pt}{1.0} 
	\small
	\begin{tabular}{l|x{36}|l|r|x{36}|x{40}|x{40}x{40}}
		\multicolumn{1}{c|}{Method}  & Data & Backbone  & Param & $T$ &  Modalities & UCF-101 & HMDB-51  \\  
		\shline
		XDC~\cite{alwassel2019self}  & K400    & R(2+1)D-18& 15.4M & 32 &   V\textbf{+A} &84.2 & 47.1 \\
		AVID~\cite{avid} & K400     & R(2+1)D-18& 15.4M & 32 &   V\textbf{+A} & 87.5 & 60.8\\ 
		GDT~\cite{patrick2020multi}    & K400    & R(2+1)D-18& 15.4M & 32 &   V\textbf{+A} &89.3 & 60.0 \\  \hline
		SpeedNet~\cite{benaim2020speednet}  & K400  & S3D-G& 9.1M  & 64  &   V  &  81.1  & 48.8 \\

		VTHCL~\cite{yang2020video}   & K400  & R-50, Slow pathway& 31.8M &  8 & V &82.1 & 49.2  \\  
		CoCLR   \cite{Han20}   & K400     & S3D-G& 9.1M & 32  &   V & {87.9} & 54.6  \\
		CoCLR Two-Stream  \cite{Han20}      & K400    & 2\x S3D-G  & 2\x9.1M & 32   & V & {90.6} &  {62.9}   \\
		\shline
		\bf{\oursshort}  & K400  &  R-50, Slow pathway& 31.8M  &  8  & V & 92.2 & 63.1 \\
		\bf{\oursshort}  & K400  & S3D-G & 9.1M &  32  & V & \bf{93.8} & \textbf{66.4}
	\end{tabular}
	\vspace{.1em}
	\caption{\textbf{Comparison to prior work on UCF101 and HMDB51.} All methods use K400 without labels. ``param'' indicates the number of parameters, $T$ inference frames used, in the backbone.  ``Modalities'' show modality used during training, where ``V'' is Visual and ``A'' is Audio input. 
	}
	\label{tab:ucf_hmdb}
\end{table*}

Table~\ref{tab:ucf_hmdb} shows the results. We experiment with two backbones: (i) the R-50, Slow pathway~\cite{feichtenhofer2019slowfast} which we used in all previous experiments, and (ii) S3D-G \cite{Xie18s3d}, a commonly used backbone for self-supervised video representation learning with downstream evaluation on UCF101 and HMDB51.

When comparing to prior work, we observe that our \textbf{\oursshort} obtains state-of-the-art performance on UCF101 and HMDB51 when using K400 as unlabeled data, outperforming the previous best approaches in self-supervised learning---{both} methods using visual (V) and audio (A) information. \textbf{\oursshort} provides better performance than \eg GDT~\cite{patrick2020multi} which is an audio-visual version of \textit{SimCLR}~\cite{Chen20}.  

In comparison to the best published vision-only approach, CoCLR~\cite{Han20}, which is a co-training variant of \textit{MoCo} \cite{he2020momentum} that uses RGB and optical-flow input in training,  \textbf{\oursshort} provides a significant performance gain of \textbf{+5.9}\% and \textbf{+11.4}\% top-1 accuracy on UCF101 and HMDB51, using the \textit{identical} backbone (S3D-G) and data, and even surpasses  \mbox{CoCLR}~\cite{Han20} by {+3.2}\% and {+3.5}\% top-1 accuracy when CoCLR is using Two-Streams of S3D-G for inference.

\paragraph{Discussion.}
We believe this is a very encouraging result. 
{In} the image classification domain, semi-supervised and self-supervised approaches compare even-handed; \eg see Table 7 in~\cite{Chen20} where \textit{a self-supervised approach} (SimCLR) \textit{outperforms all semi-supervised approaches} (\eg Pseudo-label, UDA, FixMatch).  {In contrast,} our state-of-the-art result suggests that for video understanding semi-supervised learning is a promising avenue for future research. {This is especially notable} 
given the flurry of research activity in self-supervised learning from video in this setting~\cite{alwassel2019self,avid,patrick2020multi,benaim2020speednet,Han20,yang2020video}, compared to the relative lack of past research in semi-supervised learning from video.

\section{Conclusion}\label{sec:conclusions}

This paper has presented a \ours framework that capitalizes on multiple complementary views for semi-supervised learning {for video}.
On multiple video recognition datasets, our method substantially outperforms its supervised counterpart and its semi-supervised counterpart that only considers RGB views.  
We obtain state-of-the-art performance on UCF-101 and HMD-B51 when using Kinetics-400 as unlabeled data.
In future work we plan to explore ways to automatically retrieve the most relevant unlabeled videos to assist semi-supervised video learning. 

\newcount\cvprrulercount
\appendix

\setcounter{table}{0}
\renewcommand{\thetable}{A.\arabic{table}}	

\setcounter{figure}{0}
\renewcommand{\thefigure}{A.\arabic{figure}}	
\thispagestyle{empty}

\section {Additional implementation details } \label{sec:app_details} 

All our epoch measures in the paper are based only on the \textit{labeled} data. Therefore, training 800 and 400 epochs on a 1\% and a 10\% fraction of K400 corresponds to the number of iterations that 24 and 120 epochs on 100\% of K400 would take respectively. Similarly, training 1200 and 600 epochs on a 1\% and a 10\% fraction of UCF101 corresponds to the number of iterations that 48 and 240 epochs on 100\% of UCF101 would take respectively (note we use $\mu=3$ for K400 and $\mu=4$ for UCF101, where $\mu$ is the ratio to balance the number of labeled and unlabeled data).

The learning rate is linearly annealed for the first 34 epochs~\cite{Goyal2017}. 
We follow the learning rate schedule used in~\cite{feichtenhofer2019slowfast} with 
a half-period cosine schedule \cite{Loshchilov2016}. 
In particular, the learning rate at the $n$-th iteration is $\eta\cdot0.5[\cos(\frac{n}{n_\text{max}}\pi)+1]$, where $n_\text{max}$ is the maximum training iterations and the base learning rate $\eta$ is 0.8.
We use the initialization in \cite{He2015}.  

We adopt synchronized SGD optimization in 64 GPUs following the recipe in \cite{Goyal2017}.
We train with Batch Normalization (BN) \cite{Ioffe2015}, and the BN statistics are computed within {the clips that are on the same GPU.}
We use momentum of 0.9 and SGD weight decay of 10$^\text{-4}$. Dropout \cite{srivastava2014dropout} of 0.5 is used before the final classifier layer.
The mini-batch size is 4 clips per GPU (4$\times$64${=}$256 overall) for labeled data and $4\times \mu$  clips per GPU (4$\times\mu \times$64${=}$256$\times \mu$ overall) for unlabeled data. 

In \sref{sec:result_main}, we use curriculum warm up with the following schedule. For K400, we train the model for 400, with 200 warm-up, epochs and 800, with 80 warm-up, epochs for the 10\%  and 1\% subsets respectively. For UCF101, we train the model for 600, with 80 warm-up, epochs and 1200 with no warm-up epochs for the 10\%  and 1\% subsets respectively.

{\small
\bibliographystyle{ieee_fullname}
\bibliography{egbib}

\begin{thebibliography}{10}\itemsep=-1pt

\bibitem{alwassel2019self}
Humam Alwassel, Dhruv Mahajan, Lorenzo Torresani, Bernard Ghanem, and Du Tran.
\newblock Self-supervised learning by cross-modal audio-video clustering.
\newblock {\em arXiv preprint arXiv:1911.12667}, 2019.

\bibitem{benaim2020speednet}
Sagie Benaim, Ariel Ephrat, Oran Lang, Inbar Mosseri, William~T Freeman,
  Michael Rubinstein, Michal Irani, and Tali Dekel.
\newblock Speednet: Learning the speediness in videos.
\newblock In {\em CVPR}, 2020.

\bibitem{blum1998combining}
Avrim Blum and Tom Mitchell.
\newblock Combining labeled and unlabeled data with co-training.
\newblock In {\em Proceedings of the eleventh annual conference on
  Computational learning theory}, 1998.

\bibitem{bruhn2005lucas}
Andr{\'e}s Bruhn, Joachim Weickert, and Christoph Schn{\"o}rr.
\newblock Lucas/kanade meets horn/schunck: Combining local and global optic
  flow methods.
\newblock {\em International journal of computer vision}, 2005.

\bibitem{carreira2018short}
Joao Carreira, Eric Noland, Andras Banki-Horvath, Chloe Hillier, and Andrew
  Zisserman.
\newblock A short note about kinetics-600.
\newblock {\em arXiv preprint arXiv:1808.01340}, 2018.

\bibitem{carreira2019short}
Joao Carreira, Eric Noland, Chloe Hillier, and Andrew Zisserman.
\newblock A short note on the kinetics-700 human action dataset.
\newblock {\em arXiv preprint arXiv:1907.06987}, 2019.

\bibitem{carreira2017quo}
Joao Carreira and Andrew Zisserman.
\newblock Quo vadis, action recognition? a new model and the kinetics dataset.
\newblock In {\em CVPR}, 2017.

\bibitem{Chen20}
Ting Chen, Simon Kornblith, Mohammad Norouzi, and Geoffrey Hinton.
\newblock A simple framework for contrastive learning of visual
  representations.
\newblock {\em arXiv preprint arXiv:2002.05709}, 2020.

\bibitem{chen2020big}
Ting Chen, Simon Kornblith, Kevin Swersky, Mohammad Norouzi, and Geoffrey
  Hinton.
\newblock Big self-supervised models are strong semi-supervised learners.
\newblock In {\em NeurIPS}, 2020.

\bibitem{randaug}
Ekin~D Cubuk, Barret Zoph, Jonathon Shlens, and Quoc~V Le.
\newblock Randaugment: Practical automated data augmentation with a reduced
  search space.
\newblock In {\em CVPRW}, 2020.

\bibitem{dalal2005histograms}
Navneet Dalal and Bill Triggs.
\newblock Histograms of oriented gradients for human detection.
\newblock In {\em CVPR}, volume~1, 2005.

\bibitem{dalal2006human}
Navneet Dalal, Bill Triggs, and Cordelia Schmid.
\newblock Human detection using oriented histograms of flow and appearance.
\newblock In {\em ECCV}, 2006.

\bibitem{cutout}
Terrance DeVries and Graham~W Taylor.
\newblock Improved regularization of convolutional neural networks with cutout.
\newblock {\em arXiv preprint arXiv:1708.04552}, 2017.

\bibitem{DollarPETS05}
P. Doll{\'a}r, V. Rabaud, G. Cottrell, and S. Belongie.
\newblock Behavior recognition via sparse spatio-temporal features.
\newblock In {\em Second Joint IEEE International Workshop on Visual
  Surveillance and Performance Evaluation of Tracking and Surveillance, In
  conjunction with the ICCV}, 2005.

\bibitem{fan2020pyslowfast}
Haoqi Fan, Yanghao Li, Bo Xiong, Wan-Yen Lo, and Christoph Feichtenhofer.
\newblock {PySlowFast}.
\newblock \url{https://github.com/facebookresearch/slowfast}, 2020.

\bibitem{feichtenhofer2019slowfast}
Christoph Feichtenhofer, Haoqi Fan, Jitendra Malik, and Kaiming He.
\newblock Slowfast networks for video recognition.
\newblock In {\em ICCV}, 2019.

\bibitem{feichtenhofer2016convolutional}
Christoph Feichtenhofer, Axel Pinz, and Andrew Zisserman.
\newblock Convolutional two-stream network fusion for video action recognition.
\newblock In {\em CVPR}, 2016.

\bibitem{girdhar2019distinit}
Rohit Girdhar, Du Tran, Lorenzo Torresani, and Deva Ramanan.
\newblock Distinit: Learning video representations without a single labeled
  video.
\newblock In {\em ICCV}, 2019.

\bibitem{Goyal2017}
Priya Goyal, Piotr Doll{\'a}r, Ross Girshick, Pieter Noordhuis, Lukasz
  Wesolowski, Aapo Kyrola, Andrew Tulloch, Yangqing Jia, and Kaiming He.
\newblock Accurate, large minibatch sgd: Training imagenet in 1 hour.
\newblock {\em arXiv preprint arXiv:1706.02677}, 2017.

\bibitem{goyal2017something}
Raghav Goyal, Samira~Ebrahimi Kahou, Vincent Michalski, Joanna Materzynska,
  Susanne Westphal, Heuna Kim, Valentin Haenel, Ingo Fruend, Peter Yianilos,
  Moritz Mueller-Freitag, et~al.
\newblock The" something something" video database for learning and evaluating
  visual common sense.
\newblock In {\em ICCV}, 2017.

\bibitem{grandvalet2005semi}
Yves Grandvalet and Yoshua Bengio.
\newblock Semi-supervised learning by entropy minimization.
\newblock In {\em Neurips}, 2005.

\bibitem{Han20}
Tengda Han, Weidi Xie, and Andrew Zisserman.
\newblock Self-supervised co-training for video representation learning.
\newblock In {\em Neurips}, 2020.

\bibitem{he2020momentum}
Kaiming He, Haoqi Fan, Yuxin Wu, Saining Xie, and Ross Girshick.
\newblock Momentum contrast for unsupervised visual representation learning.
\newblock In {\em CVPR}, 2020.

\bibitem{He2015}
Kaiming He, Xiangyu Zhang, Shaoqing Ren, and Jian Sun.
\newblock Delving deep into rectifiers: Surpassing human-level performance on
  imagenet classification.
\newblock 2015.

\bibitem{He16}
Kaiming He, Xiangyu Zhang, Shaoqing Ren, and Jian Sun.
\newblock Deep residual learning for image recognition.
\newblock 2016.

\bibitem{horn1981determining}
Berthold~KP Horn and Brian~G Schunck.
\newblock Determining optical flow.
\newblock In {\em Techniques and Applications of Image Understanding}, volume
  281, pages 319--331. International Society for Optics and Photonics, 1981.

\bibitem{henaff2020data}
Olivier~J. Hénaff, Aravind Srinivas, Jeffrey~De Fauw, Ali Razavi, Carl
  Doersch, S.~M.~Ali Eslami, and Aaron van~den Oord.
\newblock Data-efficient image recognition with contrastive predictive coding.
\newblock In {\em ICML}, 2020.

\bibitem{Ioffe2015}
Sergey Ioffe and Christian Szegedy.
\newblock Batch normalization: Accelerating deep network training by reducing
  internal covariate shift.
\newblock 2015.

\bibitem{videossl}
Longlong Jing, Toufiq Parag, Zhe Wu, Yingli Tian, and Hongcheng Wang.
\newblock Videossl: Semi-supervised learning for video classification.
\newblock {\em arXiv:2003.00197}, 2020.

\bibitem{kay2017kinetics}
Will Kay, Joao Carreira, Karen Simonyan, Brian Zhang, Chloe Hillier, Sudheendra
  Vijayanarasimhan, Fabio Viola, Tim Green, Trevor Back, Paul Natsev, et~al.
\newblock The kinetics human action video dataset.
\newblock {\em arXiv:1705.06950}, 2017.

\bibitem{kazakos2019epic}
Evangelos Kazakos, Arsha Nagrani, Andrew Zisserman, and Dima Damen.
\newblock Epic-fusion: Audio-visual temporal binding for egocentric action
  recognition.
\newblock In {\em ICCV}, 2019.

\bibitem{Klaeser2008}
Alexander Kl{\"a}ser, Marcin Marsza{\l}ek, and Cordelia Schmid.
\newblock A spatio-temporal descriptor based on 3d-gradients.
\newblock In {\em bmvc}, 2008.

\bibitem{korbar2018cooperative}
Bruno Korbar, Du Tran, and Lorenzo Torresani.
\newblock Cooperative learning of audio and video models from self-supervised
  synchronization.
\newblock In {\em Neurips}, 2018.

\bibitem{Kuehne11}
H. Kuehne, H. Jhuang, E. Garrote, T. Poggio, and T. Serre.
\newblock {HMDB}: A large video database for human motion recognition.
\newblock pages 2556--2563, 2011.

\bibitem{kuehne2011hmdb}
Hildegard Kuehne, Hueihan Jhuang, Est{\'\i}baliz Garrote, Tomaso Poggio, and
  Thomas Serre.
\newblock Hmdb: a large video database for human motion recognition.
\newblock In {\em ICCV}, 2011.

\bibitem{temporalensem}
Samuli Laine and Timo Aila.
\newblock Temporal ensembling for semi-supervised learning.
\newblock In {\em ICLR}, 2017.

\bibitem{lee2013pseudo}
Dong-Hyun Lee.
\newblock Pseudo-label: The simple and efficient semi-supervised learning
  method for deep neural networks.
\newblock In {\em Workshop on challenges in representation learning, ICML},
  2013.

\bibitem{liu2009beyond}
Ce Liu et~al.
\newblock {\em Beyond pixels: exploring new representations and applications
  for motion analysis}.
\newblock PhD thesis, Massachusetts Institute of Technology, 2009.

\bibitem{Loshchilov2016}
Ilya Loshchilov and Frank Hutter.
\newblock Sgdr: Stochastic gradient descent with warm restarts.
\newblock {\em arXiv preprint arXiv:1608.03983}, 2016.

\bibitem{miech2020end}
Antoine Miech, Jean-Baptiste Alayrac, Lucas Smaira, Ivan Laptev, Josef Sivic,
  and Andrew Zisserman.
\newblock End-to-end learning of visual representations from uncurated
  instructional videos.
\newblock In {\em CVPR}, 2020.

\bibitem{miyato2018virtual}
Takeru Miyato, Shin-ichi Maeda, Masanori Koyama, and Shin Ishii.
\newblock Virtual adversarial training: a regularization method for supervised
  and semi-supervised learning.
\newblock {\em IEEE transactions on pattern analysis and machine intelligence},
  2018.

\bibitem{avid}
Pedro Morgado, Vasconcelos Nuno, and Misra Ishan.
\newblock Audio-visual instance discrimination with cross-modal agreement.
\newblock {\em arXiv preprint arXiv:2004.12943}, 2020.

\bibitem{munro20multi}
Jonathan Munro and Dima Damen.
\newblock {M}ulti-modal {D}omain {A}daptation for {F}ine-grained {A}ction
  {R}ecognition.
\newblock In {\em CVPR}, 2020.

\bibitem{patrick2020multi}
Mandela Patrick, Yuki~M Asano, Ruth Fong, Jo{\~a}o~F Henriques, Geoffrey Zweig,
  and Andrea Vedaldi.
\newblock Multi-modal self-supervision from generalized data transformations.
\newblock {\em arXiv preprint arXiv:2003.04298}, 2020.

\bibitem{qian2020spatiotemporal}
Rui Qian, Tianjian Meng, Boqing Gong, Ming-Hsuan Yang, Huisheng Wang, Serge
  Belongie, and Yin Cui.
\newblock Spatiotemporal contrastive video representation learning.
\newblock {\em arXiv preprint arXiv:2008.03800}, 2020.

\bibitem{sajjadi2016regularization}
Mehdi Sajjadi, Mehran Javanmardi, and Tolga Tasdizen.
\newblock Regularization with stochastic transformations and perturbations for
  deep semi-supervised learning.
\newblock In {\em Neurips}, 2016.

\bibitem{sigurdsson2016much}
Gunnar~A Sigurdsson, Olga Russakovsky, Ali Farhadi, Ivan Laptev, and Abhinav
  Gupta.
\newblock Much ado about time: Exhaustive annotation of temporal data.
\newblock {\em arXiv preprint arXiv:1607.07429}, 2016.

\bibitem{simonyan2014two}
Karen Simonyan and Andrew Zisserman.
\newblock Two-stream convolutional networks for action recognition in videos.
\newblock In {\em Neurips}, 2014.

\bibitem{smaira2020short}
Lucas Smaira, Jo{\~a}o Carreira, Eric Noland, Ellen Clancy, Amy Wu, and Andrew
  Zisserman.
\newblock A short note on the kinetics-700-2020 human action dataset.
\newblock {\em arXiv preprint arXiv:2010.10864}, 2020.

\bibitem{fixmatch}
Kihyuk Sohn, David Berthelot, Chun-Liang Li, Zizhao Zhang, Nicholas Carlini,
  Ekin~D Cubuk, Alex Kurakin, Han Zhang, and Colin Raffel.
\newblock Fixmatch: Simplifying semi-supervised learning with consistency and
  confidence.
\newblock {\em arXiv preprint arXiv:2001.07685}, 2020.

\bibitem{soomro2012ucf101}
Khurram Soomro, Amir~Roshan Zamir, and Mubarak Shah.
\newblock Ucf101: A dataset of 101 human actions classes from videos in the
  wild.
\newblock {\em arXiv preprint arXiv:1212.0402}, 2012.

\bibitem{srivastava2014dropout}
Nitish Srivastava, Geoffrey Hinton, Alex Krizhevsky, Ilya Sutskever, and Ruslan
  Salakhutdinov.
\newblock Dropout: a simple way to prevent neural networks from overfitting.
\newblock {\em The journal of machine learning research}, 2014.

\bibitem{meanteacher}
Antti Tarvainen and Harri Valpola.
\newblock Mean teachers are better role models: Weight-averaged consistency
  targets improve semi-supervised deep learning results.
\newblock In {\em Neurips}, 2017.

\bibitem{tran2015learning}
Du Tran, Lubomir Bourdev, Rob Fergus, Lorenzo Torresani, and Manohar Paluri.
\newblock Learning spatiotemporal features with 3d convolutional networks.
\newblock In {\em ICCV}, 2015.

\bibitem{tran2018closer}
Du Tran, Heng Wang, Lorenzo Torresani, Jamie Ray, Yann LeCun, and Manohar
  Paluri.
\newblock A closer look at spatiotemporal convolutions for action recognition.
\newblock In {\em CVPR}, 2018.

\bibitem{wang2013action}
Heng Wang and Cordelia Schmid.
\newblock Action recognition with improved trajectories.
\newblock In {\em ICCV}, 2013.

\bibitem{wang2016temporal}
Limin Wang, Yuanjun Xiong, Zhe Wang, Yu Qiao, Dahua Lin, Xiaoou Tang, and Luc
  Van~Gool.
\newblock Temporal segment networks: Towards good practices for deep action
  recognition.
\newblock In {\em ECCV}, 2016.

\bibitem{xiao2020audiovisual}
Fanyi Xiao, Yong~Jae Lee, Kristen Grauman, Jitendra Malik, and Christoph
  Feichtenhofer.
\newblock Audiovisual slowfast networks for video recognition.
\newblock {\em arXiv preprint arXiv:2001.08740}, 2020.

\bibitem{uda}
Qizhe Xie, Zihang Dai, Eduard Hovy, Minh-Thang Luong, and Quoc~V Le.
\newblock Unsupervised data augmentation for consistency training.
\newblock {\em arXiv preprint arXiv:1904.12848}, 2019.

\bibitem{Xie_2020_CVPR}
Qizhe Xie, Minh-Thang Luong, Eduard Hovy, and Quoc~V. Le.
\newblock Self-training with noisy student improves imagenet classification.
\newblock In {\em CVPR}, 2020.

\bibitem{Xie18s3d}
Saining Xie, Chen Sun, Jonathan Huang, Zhuowen Tu, and Kevin Murphy.
\newblock Rethinking spatiotemporal feature learning for video understanding.
\newblock 2018.

\bibitem{xu2019self}
Dejing Xu, Jun Xiao, Zhou Zhao, Jian Shao, Di Xie, and Yueting Zhuang.
\newblock Self-supervised spatiotemporal learning via video clip order
  prediction.
\newblock In {\em CVPR}, 2019.

\bibitem{yalniz2019billion}
I~Zeki Yalniz, Herv{\'e} J{\'e}gou, Kan Chen, Manohar Paluri, and Dhruv
  Mahajan.
\newblock Billion-scale semi-supervised learning for image classification.
\newblock {\em arXiv preprint arXiv:1905.00546}, 2019.

\bibitem{yang2020video}
Ceyuan Yang, Yinghao Xu, Bo Dai, and Bolei Zhou.
\newblock Video representation learning with visual tempo consistency.
\newblock {\em arXiv preprint arXiv:2006.15489}, 2020.

\bibitem{zeng2017semi}
Ming Zeng, Tong Yu, Xiao Wang, Le~T Nguyen, Ole~J Mengshoel, and Ian Lane.
\newblock Semi-supervised convolutional neural networks for human activity
  recognition.
\newblock In {\em IEEE International Conference on Big Data}, 2017.

\bibitem{zhao2018recognize}
Yue Zhao, Yuanjun Xiong, and Dahua Lin.
\newblock Recognize actions by disentangling components of dynamics.
\newblock In {\em CVPR}, 2018.

\bibitem{zhu2020actbert}
Linchao Zhu and Yi Yang.
\newblock Actbert: Learning global-local video-text representations.
\newblock In {\em CVPR}, 2020.

\end{thebibliography}
}

\end{document}